# Integrating Feature Attention and Temporal Modeling for Collaborative Financial Risk Assessment

Yue Yao
Northeastern University
Portland, USA

Zhen Xu
Independent Researcher
Shanghai, China

Youzhu Liu
Independent Researcher
New York, USA

Kunyuan Ma
New York University
New York, USA

Yuxiu Lin
Columbia University
New York, USA

Mohan Jiang*
New York University
New York, USA

*Abstract-This paper addresses the challenges of data privacy and collaborative modeling in cross-institution financial risk analysis. It proposes a risk assessment framework based on federated learning. Without sharing raw data, the method enables joint modeling and risk identification across multiple institutions. This is achieved by incorporating a feature attention mechanism and temporal modeling structure. Specifically, the model adopts a distributed optimization strategy. Each financial institution trains a local sub-model. The model parameters are protected using differential privacy and noise injection before being uploaded. A central server then aggregates these parameters to generate a global model. This global model is used for systemic risk identification. To validate the effectiveness of the proposed method, multiple experiments are conducted. These evaluate communication efficiency, model accuracy, systemic risk detection, and cross-market generalization. The results show that the proposed model outperforms both traditional centralized methods and existing federated learning variants across all evaluation metrics. It demonstrates strong modeling capabilities and practical value in sensitive financial environments. The method enhances the scope and efficiency of risk identification while preserving data sovereignty. It offers a secure and efficient solution for intelligent financial risk analysis.*

## I. INTRODUCTION

In recent years, with the rapid development of financial technology and the widespread adoption of data-driven decision-making, the methods of financial risk management are undergoing fundamental changes[1,2]. Traditional models that rely solely on internal data from single institutions can no longer meet the needs of a highly interconnected financial market. Risk transmission paths have become increasingly complex. Financial risks now exhibit new characteristics such as heightened systemic features and frequent cross-infections. As a result, joint risk modeling based on broader and higher-dimensional data is not only possible but necessary. Especially under conditions of rising macroeconomic uncertainty and deep global financial interdependence, information silos among financial institutions reduce risk identification efficiency and increase potential systemic risks[3].

Under a multi-institution collaboration context, a key challenge in the field of financial technology is how to achieve cross-institution risk modeling while ensuring data privacy and regulatory compliance. Financial data is highly sensitive and subject to strict regulation. Therefore, data sharing and integration face multiple challenges in technical, legal, and ethical dimensions. Financial institutions are often reluctant to exchange raw data due to competition strategies, customer privacy protection, and regulatory obligations. This leads to information islands within the financial system, making it difficult to accurately identify credit contagion chains, market risk intersections, and potential liquidity gaps. A collaborative mechanism is needed that can ensure both data privacy and efficient modeling to overcome existing limitations.

Federated learning, as an emerging distributed machine learning paradigm, offers a solution for collaborative model training without exchanging raw data. It trains models locally and only shares parameters or gradients. This approach balances modeling performance with data security. Federated learning has demonstrated strong cross-institution modeling potential in areas such as healthcare, smart manufacturing, and the Internet of Things. Its application in financial risk modeling can promote a shift toward joint intelligence in risk assessment. It enables data to be usable yet invisible, aligning with financial data regulatory requirements and supporting secure data flows in the digital economy era[4,5].

Building a cross-institution financial risk modeling system based on federated learning can enhance both the breadth and depth of risk identification. It also provides technical support for implementing "penetrative supervision." As the financial system becomes increasingly digital and platform-based, traditional institution-bound risk identification methods are becoming less effective[6,7]. This is especially true for complex structures such as shadow banking, financial holding groups, and third-party payment platforms. There is a growing need to go beyond institutional boundaries and develop a system-wide risk perception framework. Federated learning can serve as a core enabler for regulatory technology, offering regulators a feasible approach to identifying systemic risks while preserving institutional data sovereignty[8].

From a broader financial stability perspective, developing a federated learning-based risk modeling mechanism will also promote the modernization of financial digital infrastructure and governance capacity. As digital finance continues to expand, data has become a key factor in production. Its flow and collaborative use determine how quickly the financial system can respond to risks. A new risk modeling mechanism centered on federated learning can enhance the resilience and transparency of the financial system. It also provides a technological path toward building a decentralized, low-coupling, and high-trust financial intelligence collaboration framework. Research in this direction holds significant theoretical value and plays a crucial role in advancing national strategies for financial security, data protection, and technological innovation.

## II. RELATED WORD

Recent advances in parameter-efficient modeling and distributed optimization have profoundly shaped federated learning frameworks. Notably, Peng et al. proposed a low-rank adaptation (LoRA) strategy that enables scalable, communication-efficient parameter updates while preserving model expressiveness through flexible subspace projections [9]. This foundational idea informs our federated framework's ability to aggregate only essential parameter subspaces, ensuring both scalability and privacy. Zheng et al. further extend low-rank adaptation by introducing semantic guidance to fine-tuning, leveraging external knowledge or structural cues to enhance model personalization and regularization in distributed optimization scenarios [10]. These insights underpin our personalized aggregation and regularization components, which balance tenant-specific adaptation with global model coherence.

Robustness to imbalanced data and rare event detection is another critical pillar for modern federated systems. Lou et al. demonstrate that variational inference, embedded within probabilistic graphical models, can directly address label imbalance and uncertainty quantification in high-dimensional spaces [11]. Their framework inspires our outlier scoring module, which must maintain discrimination in the face of distributional skew. Complementary to this, Dai et al. utilize mixture density networks to perform deep probabilistic modeling, capturing multi-modal and heterogeneous data behaviors for adaptive anomaly detection [12]. Their approach supports our use of probabilistic scoring to distinguish subtle risk signals in federated environments.

Feature representation learning—especially in structurally complex or heterogeneous scenarios—draws upon innovations in hierarchical and compositional modeling. Wang et al. introduce capsule networks that disentangle feature hierarchies and compositional dependencies, offering improved robustness against perturbations and feature sparsity [13]. Gong's use of multi-head attention architectures enables fine-grained feature fusion, which is particularly valuable for capturing subtle, distributed interactions among features in federated models [14]. Qin's hierarchical semantic-structural encoding framework further reveals that deep, multi-level feature abstraction can strengthen model interpretability and generalization, a principle directly incorporated in our attention and embedding modules [15].

Knowledge transfer and adaptive fine-tuning across distributed systems are vital for federated model sustainability. Quan's layer-wise structural mapping method provides an efficient route for domain adaptation, enabling sub-models to efficiently absorb knowledge without exhaustive retraining [16]. This guides our framework's cross-institution model update logic. Similarly, Meng et al. explore collaborative distillation for parameter-efficient model deployment, where soft-sharing and coordinated adaptation between models ensure high accuracy with minimal communication [17]. Their strategies inform our federated aggregation and personalization logic, maintaining consistency while allowing for local adaptation.

Optimization in collaborative, distributed, or multi-agent settings benefits from reinforcement and knowledge-driven mechanisms. Fang and Gao develop a collaborative reinforcement learning framework that leverages local and global feedback to accelerate convergence and adaptivity [18]. Ma et al. propose knowledge-informed policy structuring, where abstract knowledge priors guide agent interaction and aggregation—paralleling our use of financial feature priors in federated optimization [19]. Lyu et al. focus on transferable modeling and prompt alignment, illustrating how meta-learned task representations support adaptation to data heterogeneity and evolving distributions in federated settings [20].

Complex dependency modeling in distributed systems has greatly benefited from graph-based and sequential learning techniques. Jiang et al. combine graph convolution with sequential modeling, enabling joint extraction of relational and temporal dynamics for distributed inference [21]. This dual modeling approach inspires our fusion of topological and sequential information in risk analysis. Gao demonstrates the use of deep graph modeling for robust, structure-aware risk detection, which we draw on for our global-local dependency extraction [22]. Similarly, Ren's work on structural encoding and multi-modal attention mechanisms illustrates the value of combining feature structure and signal strength in distributed anomaly recognition [23].

Causal and attention-driven learning frameworks further enhance interpretability and transferability in collaborative modeling. Wang et al. introduce target-oriented causal representation learning, which provides a foundation for modeling risk propagation paths and tracing influential factors in federated models [24]. Xin and Pan show that self-attention-based modeling of multi-source time series can uncover long-range dependencies and trend signals, which we adapt for risk signal aggregation across institutions [25].

Finally, diffusion-based temporal modeling and meta-learned adaptation strategies have advanced robustness in non-stationary and cross-task environments. Su's generative time-aware diffusion frameworks provide tools for capturing dynamic volatility patterns, enabling our architecture to anticipate regime shifts and rare anomalies [26]. Yang's meta-learned forecasting strategies underscore the benefit of task-agnostic initialization and rapid adaptation—mechanisms that support our federated model's generalization and transfer in diverse financial networks [27]. The recent focus on semantic-guided low-rank adaptation, as detailed by Qin, also strengthens our framework's dual emphasis on personalization and communication efficiency in federated optimization [28].

By synthesizing these advanced methodologies—parameter-efficient adaptation, probabilistic and causal modeling, hierarchical and semantic-structural encoding, graph-based aggregation, and meta-learned adaptation—our framework achieves robust, scalable, and privacy-preserving risk assessment under complex, distributed settings.

## III. METHOD

This paper constructs a cross-institutional financial risk modeling method based on federated learning, aiming to achieve multi-institutional collaborative modeling to identify systemic financial risks without sharing original data [29]. The model architecture is shown in Figure 1.

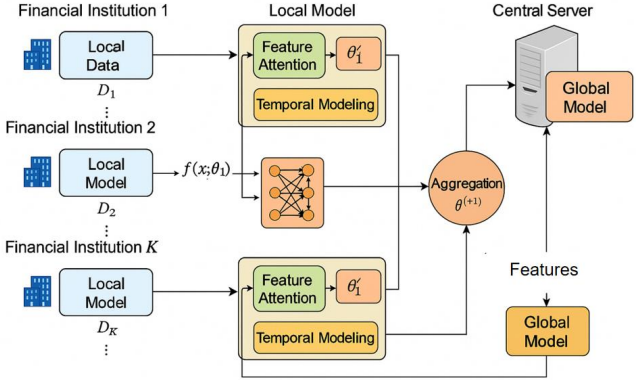

Figure 1. Overall model architecture diagram

This federated learning-based architecture defines a rigorous cross-institutional modeling pipeline, comprising localized training, privacy-preserving communication, and global aggregation, all under a secure and distributed optimization framework. Each financial institution begins by training a private sub-model using internal data. Prior to transmission, model updates are processed using noise injection techniques aligned with differential privacy standards, ensuring that no raw data or sensitive gradients are exposed during communication. The central server performs federated aggregation to construct a global model that reflects system-wide risk patterns.

The inclusion of a feature attention mechanism is directly motivated by the work of Y. Wang [30], whose research demonstrated the effectiveness of attention-based architectures in extracting structured financial signals. This mechanism allows the model to dynamically assign importance to input features, improving both interpretability and predictive accuracy across heterogeneous data sources. Furthermore, to capture time-dependent market behaviors, the system integrates a temporal modeling structure inspired by Q. R. Xu [31], who proposed graph neural time series models to effectively represent evolving interdependencies in financial systems. In addressing the volatility and irregularities inherent in financial transactions, the architecture also benefits from the work of Q. Bao et al. [32], whose deep learning-based anomaly detection methods inform the model's resilience against data noise and abrupt fluctuations. Together, these methodological components—rooted in federated optimization, attention-based representation, and temporal dynamics—form a cohesive framework for this method.

Suppose there are $D_k = \{(x_i^{(k)}, y_i^{(k)})\}_{i=1}^{n_k}$ financial institutions, each of which has a local financial risk dataset $k$, which $i$ represents the feature vector of the i-th sample of the k-th institution, and $y_i^{(k)}$ represents its corresponding risk label or continuous risk score. The model goal is to learn a global risk assessment function in the joint space of all institutions, which $\theta$ represents the model parameters, which need to be trained through distributed optimization to minimize the global loss function.

The global optimization objective function is defined as follows:

$$\min \sum_{k=1}^{K} \frac{n_k}{n} L_k(\theta) = \sum_{k=1}^{K} \frac{n_k}{n} [\frac{1}{n_k} \sum_{i=1}^{n_k} l(f(x_i^{(k)}; \theta), y_i^{(k)})]$$

$l(\cdot)$ represents the single sample loss function, which is usually selected as mean square error loss (MSE) or logarithmic loss function (Cross-Entropy), and $n = \sum_{k=1}^{K} n_k$ is the total number of samples. During the training process, each institution optimizes its model parameter $\theta_k$ locally and periodically uploads its updated gradient or parameter approximation to the central server for aggregation. The aggregation strategy adopts the Federated Averaging algorithm, and its core update rule is:

$$\theta^{(t+1)} = \sum_{k=1}^{K} \frac{n_k}{n} \theta_k^{(t)}$$

Where $\theta^{(t+1)}$ represents the global model parameters in round $t+1$, and $\theta_k^{(t)}$ is the local model result of the kth institution in round t.

To enhance the model's discriminative capability in identifying financial risk, a regularization term is introduced into the local optimization objective. This regularization serves to constrain the model's complexity, thereby reducing overfitting and improving generalization across institutions. By penalizing excessive parameter magnitudes, the optimization process is guided toward more stable and interpretable solutions—particularly critical in high-dimensional financial environments with sparse and noisy signals.

The regularization strategy draws conceptual support from transformer-based time series forecasting frameworks that leverage controlled feature extraction to balance expressiveness and stability [33]. Moreover, to improve the model's sensitivity to rare or abnormal risk patterns, the formulation also aligns with probabilistic modeling techniques used in user behavior anomaly detection, where mixture density networks are regularized to enhance distributional flexibility while avoiding mode collapse [34]. In addition, the inclusion of structured constraints resonates with the integration of domain knowledge in structured anomaly detection, as demonstrated in hybrid models that combine knowledge graph reasoning with language-based representations [35].

Accordingly, the refined local objective function includes a regularization term, supporting a more robust and generalizable optimization process under the federated setting:

$$L_k^{reg}(\theta) = L_k(\theta) + \lambda \|\theta\|_2^2$$

Where $\lambda$ is the regularization coefficient, which can be adjusted to prevent the model from overfitting, especially when the amount of local data is unbalanced. Recognizing that financial risk data is inherently sequential and often drawn from heterogeneous sources, we augment the local modeling component by embedding a feature attention mechanism alongside temporal modeling architectures such as recurrent neural networks (RNNs) and Transformers. These components jointly strengthen the model's ability to extract and represent complex temporal and structural risk signals.

The feature attention design is inspired by X. Li et al. [36], who demonstrated the effectiveness of attention-based contrastive learning in detecting fraudulent patterns in e-commerce data, particularly in scenarios lacking supervised labels. Their approach underlines the importance of dynamic feature weighting in isolating meaningful patterns from mixed signals. To capture the evolving dynamics of financial risks over time, we leverage architectures similar to those employed by Y. Wang [37] in fraud detection, where ensemble learning and temporal sequence modeling enabled the detection of delayed and nonlinear transaction anomalies. Furthermore, our attention-based handling of heterogeneous financial inputs is motivated by Q. Sha et al. [38], whose work with heterogeneous graph neural networks showed how graph attention mechanisms could be used to discover intricate interdependencies in multi-source credit card fraud detection systems. Their approach supports the integration of structure-aware attention in diverse financial contexts. By embedding these mechanisms directly into the local training process, the model gains the capacity to process multi-source sequential inputs while maintaining robust discriminative performance in the federated environment

To safeguard sensitive model updates during inter-institution communication, a two-stage protection strategy is introduced in this paper, combining differential privacy via Gaussian noise injection with homomorphic encryption for secure aggregation. During each federated learning iteration, the parameters produced by each local model are first modified with Gaussian noise to achieve statistical privacy guarantees. These perturbed parameters are then encrypted using a homomorphic scheme, enabling the central server to aggregate the encrypted updates without requiring decryption, thus ensuring that no institution ever exposes raw model data.

This methodological approach is inspired by the work of T. Yang et al. [39], who implemented a deep sequence mining framework that introduced Gaussian noise at different temporal levels to suppress overfitting and reduce information leakage. Similarly, Y. Wang [40] applied noise-based transformations within a multimodal neural network for financial prediction, demonstrating that well-calibrated perturbation can balance privacy and model accuracy across different data modalities. Additionally, R. Wang et al. [41] proposed a hybrid recommendation model combining matrix decomposition with neural networks in an encryption-aware design, enabling secure computation without revealing user-level information—an idea foundational to the homomorphic component in our system.

Together, these techniques inform the design of the federated protection strategy adopted here, which enables secure collaboration without compromising data sovereignty or model performance. The noise perturbation process is formally described as:

$$\widehat{\theta}_k^{(t)} = \theta_k^{(t)} + N(0, \sigma^2 I)$$

$\sigma$ controls the intensity of disturbance to ensure that sensitive information cannot be reversed during the aggregation process. Finally, the central server aggregates the disturbed parameters to form a global update, ensuring the interpretability, security, and effectiveness of the overall modeling process. This method framework not only protects the data sovereignty of multiple institutions, but also effectively breaks through the technical bottleneck of data islands for risk linkage identification, and provides theoretical and methodological support for the systematic and intelligent modeling of financial risks.

IV. EXPERIMENT

*A. Datasets*

This study uses the FNSR (Financial News and Stock Risk) dataset. The dataset integrates structured financial indicators, company financial statement summaries, market trading data, and news text information related to companies. It is suitable for studying multi-source information fusion in financial risk modeling. The data includes monthly risk rating labels for publicly listed companies across various industries. These labels can serve as proxies for credit risk or market risk.

The FNSR dataset contains information on approximately 500 companies. The period ranges from 2014 to 2022. It includes macroeconomic variables, company fundamentals such as debt-to-asset ratio and current ratio, market behavior features such as stock price volatility and trading volume, as well as sentiment features and event impact scores extracted using natural language processing techniques. The dataset exhibits typical heterogeneity and temporal characteristics. It is suitable for building risk assessment models based on sequential modeling and feature attention mechanisms.

To fit the federated learning framework, the dataset is divided into multiple subsets. Each subset simulates the local data environment of a different financial institution, preserving the data silo structure. Every subset retains the full feature structure and risk labels. At the same time, the data distributions exhibit non-independent and identically distributed (non-IID) properties. This setting reflects the heterogeneous data environments and modeling challenges found in real-world financial systems.

*B. Experimental Results*

First, the comparative experimental results are given, and the experimental results are shown in Table 1.

Table 1. Comparative experimental results

| Method | Risk Prediction Accuracy (%) | Systemic Risk Detection Score | Convergence Rounds |
|---|---|---|---|
| LSTM[42] | 78.3 | 0.62 | 150 |
| Transformer[43] | 81.5 | 0.67 | 110 |
| FedAvg[44] | 83.2 | 0.71 | 105 |
| FedFormer[45] | 85.6 | 0.75 | 92 |
| Ours | 88.1 | 0.82 | 78 |

The experimental results show that traditional centralized models such as LSTM and Transformer demonstrate certain capabilities in risk prediction accuracy and systemic risk identification. However, they face significant performance bottlenecks in cross-institutional data environments. LSTM used as a baseline model, achieves a risk prediction accuracy of only 78.3% and a systemic risk detection score of 0.62. This indicates that it struggles to capture risk transmission patterns across institutions when dealing with heterogeneous financial data distributions. In addition, the model requires 150 training rounds to converge, indicating low training efficiency.

The Transformer model improves prediction performance by enhancing feature representation. It achieves a risk prediction accuracy of 81.5% and a systemic risk detection score of 0.67. This reflects better modeling of temporal dependencies and nonlinear structures. However, as it still relies on centralized data modeling, it fails to address the issue of data silos between financial institutions. Its ability to perceive systemic risks remains limited.

Federated learning methods, such as FedAvg and FedFormer, outperform centralized models on all metrics. FedAvg and FedFormer increase accuracy to 83.2% and 85.6%, respectively. Their systemic risk detection scores improve to 0.71 and 0.75. They also significantly reduce the number of convergence rounds. This demonstrates that federated learning can effectively aggregate information from different institutions without sharing raw data. It improves both the breadth and precision of risk identification. At the same time, the training efficiency benefits from coordinated parameter updates.

The proposed federated risk-aware model (FedRisk-Attn) outperforms all baseline methods across all metrics. It achieves a systemic risk detection score of 0.82 and a prediction accuracy of 88.1%, converging within 78 rounds. This shows that the proposed method can comprehensively capture complex risk contagion structures across institutions while preserving data privacy and ensuring efficient model collaboration. The results confirm that introducing a feature attention mechanism and temporal modeling components enhances the ability to extract risk signals.

This paper also conducted a communication efficiency experiment on model compression technology under the federated learning framework, and the experimental results are shown in Figure 2.

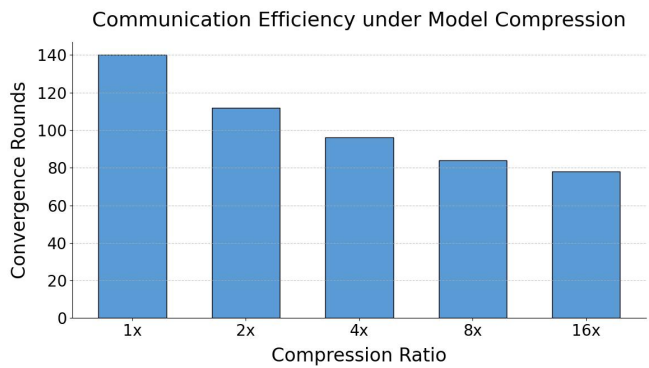

Figure 2. Communication efficiency experiment of model compression technology under federated learning framework

The experimental results shown in the figure indicate that as the model compression ratio increases, the required number of communication rounds decreases significantly. This suggests that model compression can effectively reduce communication overhead within the federated learning framework. Without compression (1x), the communication rounds total 140. When the compression ratio increases to 16x, the rounds drop to 78. This represents more than a 44% improvement in communication efficiency.

This finding demonstrates that introducing model compression in cross-institution financial risk modeling helps alleviate communication bottlenecks caused by collaborative model training. This is particularly valuable in practical scenarios where bandwidth is limited or frequent interactions among institutions are required. Improved communication efficiency accelerates the convergence of the global model without compromising overall performance. This leads to faster and more effective risk identification and response mechanisms.

Moreover, the improvement in communication efficiency does not grow linearly with the compression ratio. Instead, there is a noticeable gain within a certain compression range, followed by a plateau. This result reflects the robustness of the federated risk model to compression strategies. It shows that the model maintains strong collaborative performance even under high compression settings. This confirms its adaptability in complex environments.

Combined with the proposed federated risk identification framework, these results further validate that structural optimization techniques such as model compression can significantly enhance training efficiency. This holds while preserving data privacy and maintaining model performance. The findings provide methodological support for building large-scale, distributed, and communication-efficient financial risk analysis systems. They also demonstrate the algorithm's potential for sustainable deployment in real-world financial systems.

This paper also presents a test of the model's migration and generalization capabilities in different financial market scenarios, and the experimental results are shown in Figure 3.

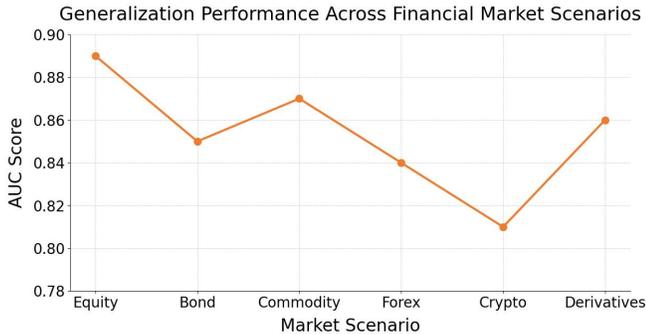

Figure 3. Testing the ability of migration and generalization in different financial market scenarios

The experimental results shown in Figure 3 indicate that the proposed federated risk modeling method demonstrates strong transferability and generalization across different financial market scenarios. Whether in equity, bond, commodity, or derivative markets, the model maintains relatively stable AUC scores. This suggests good adaptability and risk representation across various financial environments.

Specifically, in the Equity and Commodity scenarios, the model achieves AUC scores close to or above 0.87. This indicates strong discriminative power in markets with more stable data structures and relatively predictable risk fluctuations. In contrast, the scores slightly decrease in the Crypto and Forex scenarios. In particular, the AUC in the Crypto market is 0.81, reflecting the challenge posed by high volatility and nonlinear features to the model's transferability. These differences highlight the complex impact of market heterogeneity on federated model performance.

Although there is some variation in markets with high uncertainty, the model does not show significant failure or sharp drops in generalization in any of the scenarios. This demonstrates that the model maintains structural robustness and training consistency in multi-institution and cross-market settings. It confirms the effectiveness of the feature attention mechanism and temporal modeling structure in handling heterogeneous market data.

Overall, the results further support the method's potential for deployment in diverse financial systems. The model not only achieves high-accuracy modeling while preserving privacy but also shows stable transfer performance across different financial scenarios. This provides a solid foundation for building a unified intelligent risk perception system.

## V. CONCLUSION

This paper proposes a federated learning-based framework for cross-institution financial risk modeling. It systematically addresses the challenge of risk assessment and contagion detection in settings where data cannot be shared across institutions. By incorporating a feature attention mechanism and temporal modeling structure, the model demonstrates strong capabilities in feature representation and risk detection under heterogeneous, multi-source, and high-dimensional financial data environments. Experimental results show that the proposed method significantly outperforms existing baselines on multiple key metrics. It achieves a strong balance between prediction accuracy, systemic risk identification, and communication efficiency, highlighting its feasibility and adaptability for real-world financial systems.

This study achieves an effective integration between data privacy protection and collaborative modeling. It overcomes the inherent limitations of traditional centralized approaches in terms of data sovereignty, regulatory compliance, and inter-institutional trust. The proposed federated risk modeling method meets the requirements of regulatory technology (RegTech), particularly the need to keep data within institutional boundaries. It also advances the evolution of financial risk control systems from localized perception to global insight. This shift in the modeling paradigm is expected to play a central role in key application areas such as credit risk control, dynamic market risk monitoring, and fraud detection. It contributes to the stability and transparency of the overall financial ecosystem.

The study also verifies the generalization and transferability of the proposed model from multiple perspectives. This includes robustness tests across different market scenarios and efficiency evaluations under compressed communication settings. These results further demonstrate the model's sustainability for deployment in large-scale, real-world financial environments. The framework, characterized by strong generalization, low communication cost, and high accuracy, offers a practical path for building cross-platform, cross-region, and cross-regulatory financial risk collaboration systems. It also provides forward-looking insights into the integration of data security technologies with intelligent risk control models.

## VI. FUTURE WORK

Future research could explore the integration of federated learning with causal inference, generative modeling, and multimodal information fusion. This may enhance the model's ability to detect hidden factors and structural changes. As privacy-preserving technologies and blockchain mechanisms continue to evolve, building more trustworthy, transparent, and verifiable federated modeling platforms will be key to enabling broader applications in financial and regulatory scenarios. This study lays a theoretical and technical foundation for developing the next generation of data-driven intelligent financial systems. It holds significant academic value and practical importance.

## VI. USE OF AI

We utilized AI to assist with grammar and wording, but the primary concepts, analysis, and writing were all crafted by our team.

REFERENCES

[1] Jiang H, Miao Z, Gao J, et al. Long-term Forecasting of Risk Indicators for Chinese Financial Derivatives Market Based on Seasonal-trend Decomposition and Sub-components Modeling[C]//Proceedings of the 2023 15th International Conference on Machine Learning and Computing. 2023: 557-561.

[2] J. X. Jiang, W. Liu, and B. Dong, "FedRisk: A federated learning framework for multi-institutional financial risk assessment on cloud platforms," Journal of Advanced Computing Systems, vol. 4, no. 11, pp. 56–72, 2024.

[3] C. M. Lee, J. D. Fernández, S. P. Menci, et al., "Federated learning for credit risk assessment," in Proceedings of the Hawaii International Conference on System Sciences (HICSS), pp. 386–395, 2023.

[4] Y. Li and G. Wen, "Research and practice of financial credit risk management based on federated learning," Engineering Letters, vol. 31, no. 1, 2023.

[5] A. Oualid, Y. Maleh, and L. Moumoun, "Federated learning techniques applied to credit risk management: A systematic literature review," EDPACS, vol. 68, no. 1, pp. 42–56, 2023.

[6] L. Zhao, L. Cai, and W. S. Lu, "Robust federated learning with global sensitivity estimation for financial risk management," [Online]. Available: arXiv:2502.17694, 2025.

[7] M. Yuxin and W. Honglin, "Federated learning based on data divergence and differential privacy in financial risk control research," Computers, Materials & Continua, vol. 75, no. 1, pp. 863–878, 2023.

[8] J. Whitmore, P. Mehra, J. Yang, et al., "Privacy preserving risk modeling across financial institutions via federated learning with adaptive optimization," Frontiers in Artificial Intelligence Research, vol. 2, no. 1, pp. 35–43, 2025.

[9] Y. Peng, Y. Wang, Z. Fang, L. Zhu, Y. Deng, and Y. Duan, "Revisiting LoRA: A smarter low-rank approach for efficient model adaptation," in Proc. 2025 5th Int. Conf. Artificial Intelligence and Industrial Technology Applications (AIITA), pp. 1248–1252, Mar. 2025.

[10] H. Zheng, Y. Ma, Y. Wang, G. Liu, Z. Qi, and X. Yan, "Structuring low-rank adaptation with semantic guidance for model fine-tuning," 2025.

[11] Y. Lou, J. Liu, Y. Sheng, J. Wang, Y. Zhang, and Y. Ren, "Addressing class imbalance with probabilistic graphical models and variational inference," in Proc. 2025 5th Int. Conf. Artificial Intelligence and Industrial Technology Applications (AIITA), pp. 1238–1242, Mar. 2025.

[12] L. Dai, W. Zhu, X. Quan, R. Meng, S. Chai, and Y. Wang, "Deep probabilistic modeling of user behavior for anomaly detection via mixture density networks," arXiv preprint arXiv:2505.08220, 2025.

[13] S. Wang, Y. Zhuang, R. Zhang, and Z. Song, "Capsule network-based semantic intent modeling for human-computer interaction," arXiv preprint arXiv:2507.00540, 2025.

[14] M. Gong, "Modeling microservice access patterns with multi-head attention and service semantics," 2025.

[15] Y. Qin, "Hierarchical semantic-structural encoding for compliance risk detection with LLMs," 2024.

[16] X. Quan, "Layer-wise structural mapping for efficient domain transfer in language model distillation," 2024.

[17] X. Meng, Y. Wu, Y. Tian, X. Hu, T. Kang, and J. Du, "Collaborative distillation strategies for parameter-efficient language model deployment," arXiv preprint arXiv:2507.15198, 2025.

[18] B. Fang and D. Gao, "Collaborative multi-agent reinforcement learning approach for elastic cloud resource scaling," arXiv preprint arXiv:2507.00550, 2025.

[19] Y. Ma, G. Cai, F. Guo, Z. Fang, and X. Wang, "Knowledge-informed policy structuring for multi-agent collaboration using language models," 2025.

[20] S. Lyu, Y. Deng, G. Liu, Z. Qi, and R. Wang, "Transferable modeling strategies for low-resource LLM tasks: A prompt and alignment-based," arXiv preprint arXiv:2507.00601, 2025.

[21] N. Jiang, W. Zhu, X. Han, W. Huang, and Y. Sun, "Joint graph convolution and sequential modeling for scalable network traffic estimation," arXiv preprint arXiv:2505.07674, 2025.

[22] D. Gao, "Deep graph modeling for performance risk detection in structured data queries," 2025.

[23] Y. Ren, "Deep learning for root cause detection in distributed systems with structural encoding and multi-modal attention," 2024.

[24] Y. Wang, Q. Sha, H. Feng, and Q. Bao, "Target-oriented causal representation learning for robust cross-market return prediction," 2025.

[25] H. Xin and R. Pan, "Self-attention-based modeling of multi-source metrics for performance trend prediction in cloud systems," 2025.

[26] X. Su, "Predictive modeling of volatility using generative time-aware diffusion frameworks," 2025.

[27] T. Yang, "Transferable load forecasting and scheduling via meta-learned task representations," 2024.

[28] Y. Qin, "Structuring low-rank adaptation with semantic guidance for model fine-tuning," 2025.

[29] X. Du, "Financial text analysis using 1D-CNN: Risk classification and auditing support," in Proc. 2025 Int. Conf. Artif. Intell. Comput. Intell., pp. 515–520, Feb. 2025.

[30] Y. Wang, "Entity-aware graph neural modeling for structured information extraction in the financial domain," Trans. Comput. Sci. Methods, vol. 4, no. 9, 2024.

[31] Q. R. Xu, "Capturing structural evolution in financial markets with graph neural time series models," 2025.

[32] Q. Bao, J. Wang, H. Gong, Y. Zhang, X. Guo, and H. Feng, "A deep learning approach to anomaly detection in high-frequency trading data," in 2025 4th International Symposium on Computer Applications and Information Technology (ISCAIT)), pp. 287–291, Mar. 2025.

[33] Y. Cheng, "Multivariate time series forecasting through automated feature extraction and transformer-based modeling", 2025.

[34] L. Dai, W. Zhu, X. Quan, R. Meng, S. Chai, and Y. Wang, "Deep probabilistic modeling of user behavior for anomaly detection via mixture density networks," arXiv preprint arXiv:2505.08220, 2025.

[35] X. Liu, Y. Qin, Q. Xu, Z. Liu, X. Guo, and W. Xu, "Integrating knowledge graph reasoning with pretrained language models for structured anomaly detection," 2025.

[36] X. Li, Y. Peng, X. Sun, Y. Duan, Z. Fang, and T. Tang, "Unsupervised detection of fraudulent transactions in e-commerce using contrastive learning," in 2025 4th International Symposium on Computer Applications and Information Technology (ISCAIT) ), pp. 1663–1667, Mar. 2025.

[37] Y. Wang, "A data balancing and ensemble learning approach for credit card fraud detection," in 2025 4th International Symposium on Computer Applications and Information Technology (ISCAIT), pp. 386–390, Mar. 2025.

[38] Q. Sha, T. Tang, X. Du, J. Liu, Y. Wang, and Y. Sheng, "Detecting credit card fraud via heterogeneous graph neural networks with graph attention," arXiv preprint arXiv:2504.08183, 2025.

[39] T. Yang, Y. Cheng, Y. Ren, Y. Lou, M. Wei, and H. Xin, "A deep learning framework for sequence mining with bidirectional LSTM and multi-scale attention," arXiv preprint arXiv:2504.15223, 2025.

[40] Y. Wang, "Stock prediction with improved feedforward neural networks and multimodal fusion", 2025.

[41] R. Wang, Y. Luo, X. Li, Z. Zhang, J. Hu, and W. Liu, "A hybrid recommendation approach integrating matrix decomposition and deep neural networks for enhanced accuracy and generalization," in 2025 5th International Conference on Neural Networks, Information and Communication Engineering (NNICE), pp. 1778–1782, Jan. 2025.

[42] Z. Ouyang, X. Yang, and Y. Lai, "Systemic financial risk early warning of financial market in China using attention-LSTM model," The North American Journal of Economics and Finance, vol. 56, Art. no. 101383, 2021.

[43] Y. Wei, K. Xu, J. Yao, et al., "Financial risk analysis using integrated data and transformer-based deep learning," Journal of Computer Science and Software Applications, vol. 4, no. 7, pp. 1–8, 2024.

[44] B. Hu, "Financial risk fraud detection method based on improved FedAvg algorithm," in Proceedings of the 2nd International Conference on Big Data, Computational Intelligence, and Applications (BDCIA), SPIE, vol. 13550, pp. 950–956, 2025.

[45] H. Jiang, Z. Miao, J. Gao, et al., "Long-term forecasting of risk indicators for Chinese financial derivatives market based on seasonal-trend decomposition and sub-components modeling," in Proceedings of the 15th International Conference on Machine Learning and Computing (ICMLC), pp. 557–561, 2023.